\documentclass{article}

\usepackage{times}
\usepackage{graphicx} 
\usepackage{subfigure} 

\usepackage{natbib}

\usepackage{algorithm}
\usepackage{algorithmic}

\usepackage{amsmath, amsthm, amssymb}

\usepackage{hyperref}

\usepackage[accepted]{icml2017}

\icmltitlerunning{PixelCNN Models with Auxiliary Variables for Natural Image Modeling}

\newcommand{\GRAYNAME}{\emph{Grayscale PixelCNN}}
\newcommand{\GRAY}{\GRAYNAME\xspace}
\newcommand{\GRAYNAMEs}{\emph{Grayscale PixelCNNs}}
\newcommand{\GRAYs}{\GRAYNAMEs\xspace}

\newcommand{\PYRAMIDNAME}{\emph{Pyramid PixelCNN}}
\newcommand{\PYRAMID}{\PYRAMIDNAME\xspace}
\newcommand{\PYRAMIDNAMEs}{\emph{Pyramid PixelCNNs}}
\newcommand{\PYRAMIDs}{\PYRAMIDNAMEs\xspace}

\newcommand{\SAMPLING}{MAP sampling\xspace}

\newcommand{\argmax}{\operatorname*{argmax}}

\usepackage{xspace}
\makeatletter
\DeclareRobustCommand\onedot{\futurelet\@let@token\@onedot}
\def\@onedot{\ifx\@let@token.\else.\null\fi\xspace}

\def\eg{{e.g}\onedot} 
\def\ie{{i.e}\onedot} 
 
\def\etc{{etc}\onedot}

\makeatother

\begin{document} 

\twocolumn[
\icmltitle{PixelCNN Models with Auxiliary Variables for Natural Image Modeling}




\begin{icmlauthorlist}
\icmlauthor{Alexander Kolesnikov}{ist}
\icmlauthor{Christoph H. Lampert}{ist}
\end{icmlauthorlist}

\icmlaffiliation{ist}{IST Austria, Klosterneuburg, Austria}

\icmlcorrespondingauthor{Alexander Kolesnikov}{akolesnikov@ist.ac.at}

\icmlkeywords{Image modeling, Probabilistic models,
              Unsupervised learning, Deep learning}

\vskip 0.3in
]



\printAffiliationsAndNotice{}  

\begin{abstract} 
We study probabilistic models of natural images and extend
the autoregressive family of PixelCNN architectures by
incorporating auxiliary variables.
Subsequently, we describe two new generative image models that exploit 
different image transformations as auxiliary variables: a quantized
grayscale view of the image or a multi-resolution image pyramid. 
The proposed models tackle two known shortcomings 
of existing PixelCNN models: 1) their tendency to focus on low-level image details,
while largely ignoring high-level image information, such as object shapes,
and 2) their computationally costly procedure for image sampling. 
We experimentally demonstrate benefits of the proposed models,
in particular showing that they produce much more realistically 
looking image samples than previous state-of-the-art probabilistic models.
\end{abstract} 

\section{Introduction}

Natural images are the main input to visual processing systems and,
thus, understanding their structure is important for building
strong and accurate automatic vision systems. 
Image modeling is also useful for a wide variety of key computer 
vision tasks, such as visual representation learning, 
image inpainting, deblurring, super-resolution, image compression, and
others.

Natural image modeling is known to be a very challenging
statistical problem.
Because the distribution over natural images is highly complex,
developing models that are both accurate and computationally
tractable is very challenging.
Until recently, most of the existing models were restricted to modeling 
very small image patches, no bigger than, \eg, 9x9 pixels. 
Recently, however, deep convolutional neural networks (CNNs) have 
triggered noticeable advances in probabilistic image modeling. 
Out of these, PixelCNN-type models~\cite{van2016conditional,oord2016pixel,salimans2016pixel}, 
have shown to deliver the best performance, 
while at the same time staying computationally tractable. 
However, PixelCNNs also have noticeable shortcomings:
unless conditioned on external input, the samples they produce rarely reflect
global structure of complex natural images, see~\cite{oord2016pixel,salimans2016pixel}.
This raises concerns that current PixelCNN architectures might also be  
more limited than originally presumed. 
Moreover, PixelCNN's image sampling procedure is relatively slow in practice, 
as it requires to invoke a very deep neural network for every 
single image pixel that is to be generated. 

In this work we derive improved PixelCNN models 
that address several of the aforementioned shortcomings. 
The main idea is to augment PixelCNN with appropriate auxiliary variables 
in order to isolate and overcome these drawbacks.
This step, at the same time, provides us with important insights into 
the task of modeling natural images.

Besides the above insight, we make two main technical contributions in this paper.
First, we show that uncertainty in low-level image details, such as
texture patterns, dominates the objective of ordinary probabilistic 
PixelCNN models and, thus, these models may have little incentive to capture 
visually essential high-level image information, such as object shapes.
We tackle this issue by deriving \GRAYs that effectively 
decouple the tasks of modeling low and high-level image details.
This results in image samples of substantially improved visual quality.
Second, we show that the sampling speed of PixelCNN models can be 
largely accelerated.
We accomplish this by deriving \PYRAMIDs that decompose the modeling 
of the image pixel probabilities into a series of much simpler steps. 
Employing a much lighter-weight PixelCNN architecture for each of 
them, globally coherent high-resolution samples can be obtained 
at reduced computational cost.

\section{Related Work}

Probabilistic image modeling is a research area of long tradition 
that has attracted interest from many different 
disciplines~\cite{ruderman1994statistics,olshausen1996natural,hyvarinen2009natural}.
However, because of the difficulty of the problem, until recently 
all existing models were restricted to small image patches, typically 
between 3x3 and 9x9 pixels, and reflected only low-order statistics, 
such as edge frequency and orientation~\cite{zhu1998filters,roth2005fields,carlsson2008local,zoran2011learning}. 
Utilized as prior probabilities in combination with other probabilistic 
models, such as Markov random fields~\cite{geman1984stochastic},
these models proved to be useful for low-level imaging tasks, such 
as image denoising and inpainting. 
Their expressive power is limited, though, as one can see from the fact 
that samples from the modeled distribution do not resemble natural images, 
but rather structured noise. 

This situation changed with the development of probabilistic image models 
based on deep neural networks, in particular variational auto-encoders (VAEs) 
and PixelCNNs.
In this paper, we concentrate on the latter.
The PixelCNN family of models~\cite{van2016conditional,oord2016pixel,salimans2016pixel}
factorizes the distribution of a natural image using the elementary chain 
rule over pixels. The factors are modeled as deep convolutional neural networks 
with shared parameters and trained by maximum likelihood estimation. 

VAEs~\cite{kingma2013auto} offer an alternative 
approach to probabilistic image modeling. 
They rely on a variational inequality to bound the intractable true 
likelihood of an image by a tractable approximation. 
VAEs are efficient to evaluate, but so far, produce results slightly 
worse than state-of-the-art PixelCNNs, both the likelihood scores and sampled images.
Recent advances~\cite{gregor2015draw,gregor2016towards,
                      bachman2016architecture,kingma2016improving,gulrajani2016pixelvae,
                      chen2016variational}
in VAEs literature exploit modifications of model structure, including usage of latent variables and autoregression principle,
though these techniques remain technically and conceptually different from PixelCNNs.

Specifically for the task of producing images and other complex 
high-dimensional objects, \emph{generative adversarial networks} 
(GANs)~\cite{goodfellow2014generative} have recently gained popularity.
In contrast to PixelCNNs and VAEs, GANs are not explicit probabilistic models
but feed-forward networks that are directly trained to produce naturally 
looking images from random inputs.
A drawback of GAN models is that they have a generally unstable training 
procedure, associated with the search of a Nash equilibrium between two 
competing network players, and they can suffer from various technical 
problems, such as mode collapse or vanishing gradients.
In order to make GANs work in practice, researchers resort to multiple 
non-trivial heuristics~\cite{salimans2016improved}.
This strongly contrasts with probabilistic autoregressive models,
such as PixelCNNs, which rely on well understood likelihood maximization
for training and do not suffer from mode collapse problems.

Despite having fundamental differences on the technical level, PixelCNNs, 
VAEs and GANs may also benefit each other by sharing ideas.
In particular, our work is related to the line of work on GANs with 
controllable image structure~\cite{reed2016learning,wang2016generative}.
A crucial difference of our work is, however, that our models 
do not require external supervision and, thus, remain purely unsupervised.
Another notable paper in the context of this work introduces 
Laplacian GANs~\cite{denton2015deep}, with which we share the 
similar idea of using multi-scale decomposition for image generation.
Similar constructions were suggested in \cite{oord2016pixel} in the
context of recurrent networks and \cite{dahl2017pixel} for the problem 
of super-resolution.

Very recent work~\cite{reed2017parallel}, which appeared in parallel 
with ours, also addresses PixelCNN's sampling speed with multi-scale 
decomposition. Unlike our work, this paper makes strong additional 
independence assumption on the pixel level. Our paper makes a largely 
complementary contribution, as we explore different angles in which 
a multi-scale approach can improve performance.

\section{PixelCNNs with Auxiliary Variables}\label{sec:method}

In this section we remind the reader of the 
technical background and develop a framework for
PixelCNNs with auxiliary variables.
We then propose two new PixelCNN instances that provide
insights into the natural image modeling task and lead 
to improved quality of sampled images and accelerated sampling procedure.
Finally, we conclude the section with implementation and training details.

We define an image $X = (x_1, x_2, \dots, x_{n})$ as a collection of $n$
random variables associated with some unknown probability measure $p(X)$.
Each random variable represents a 3-channel pixel value in the RGB format,
where each channel takes a discrete value from the set $\{0, 1, 2, \dots, 255\}$.
The pixels are ordered according to a raster scan order: from left to
right and from top to bottom. 
Given a dataset $\mathcal{D}$ of $N$ images, our main goal is to 
estimate the unknown probability measure $p(X)$ from $\mathcal{D}$.

Recall that PixelCNNs are a family of models~\cite{van2016conditional,oord2016pixel,salimans2016pixel}
that factorize the distribution of natural images using the basic 
chain rule:
\begin{equation}
        p(X) = \prod_{j=1}^{n} p \left(x_j | x_1, \dots, x_{j - 1} \right).
        \label{eq:originalfactorization}
\end{equation}

Our key idea is to introduce an additional auxiliary variable, $\widehat{X}$,
into the image modeling process.
Formally, a PixelCNN with auxiliary variable is a probabilistic model of the
joint distribution of $X$  and $\widehat{X}$, factorized as
\begin{equation}
        p(X, \widehat{X}) = p_{\hat{\theta}}(\widehat{X}) p_{\theta}(X | \widehat{X}),
        \label{eq:likelihood}
\end{equation}
where $[\theta, \hat{\theta}]$ are the model parameters.
The conditional probability distribution, $p_{\theta}(X|\widehat{X})$, is modeled 
using the PixelCNN model \cite{oord2016pixel}, including the
recent improvements suggested in \cite{salimans2016pixel}:
\begin{equation}
        p_{\theta}(X | \widehat{X}) = \prod_{j=1}^{n} p_{\theta}\!\left(x_j | x_1, \dots, x_{j - 1}; f_w(\widehat{X}) \right),
\end{equation}
where $f_w$ is an embedding function parametrized by a parameter vector $w$.

Like other PixelCNN-based models, our model can be used for drawing 
samples. For a fixed $\hat X$, one follows the ordinary
PixelCNN's sampling strategy.
Otherwise, one first samples $\widehat{X}$ from $p_{\hat{\theta}}(\widehat{X})$
and then samples $X$ from $p_{\theta}(X|\widehat{X})$ as described before.
Specifically, in this work we concentrate on auxiliary variables that 1) are a form 
of images themselves such that we can model $p_{\hat\theta}(\widehat{X})$  by a PixelCNN-type
model, and 2) for which $\widehat{X}$ is approximately computable by a known deterministic
function $\psi: X \rightarrow \widehat{X}$.
This choice has the useful consequence that $p(\widehat{X}|X)$ is going to be a highly
peaked distribution around the location $\psi(X)$,
which provides us with a very efficient training procedure:
denoting the model parameters $\Theta = (\hat\theta,\theta,w)$, we jointly 
maximize the log-likelihood of the observed training data 
$\mathcal{D}$ and the corresponding auxiliary variables, $\widehat X = \psi(X)$.
More precisely, we solve 
\begin{align}
\Theta^\ast &= \argmax_{\Theta} \sum_{X\in\mathcal{D}} \log  p(X, \widehat X) 
\\
&= 
\argmax_{\Theta} \sum_{X\in\mathcal{D}} \log  p_{\theta}(X|\psi(X)) + \sum_{X\in\mathcal{D}} \log  p_{\hat\theta}(\psi(X)).
\notag
\end{align}
Because the objective function decomposes and the parameters do not interact, 
we can perform the optimization over $\hat\theta$ and $(\theta, w)$ 
separately, and potentially in parallel.

Note, that by this procedure we maximize a lower bound on the log-likelihood of the data:
\begin{equation}
        \log \prod\limits_{X \in \mathcal{D}} p(X, \widehat X)
        \leq \log \prod\limits_{X \in \mathcal{D}} p(X)
        \label{eq:loglikelihood-lowerbound}
\end{equation}
By making the lower bound high we also guarantee that the log-likelihood of the data itself is high.
Furthermore, we expect the bound \eqref{eq:loglikelihood-lowerbound} to be (almost) tight, as $p(X,\widehat{X}) = p(\widehat{X}|X)p(X)$, 
and by construction the first factor, $p(\widehat{X}|X)$ 
is (almost) a $\delta$-peak centered at $\widehat{X}=\psi(X)$. 

In the rest of this section we present two concrete realizations of 
the PixelCNNs augmented with auxiliary variables.

\begin{figure}[t]
	\center
	\includegraphics[width=0.45\textwidth,trim={0cm 0cm 0cm 0cm},clip]{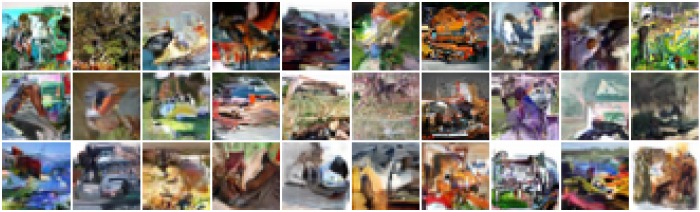}

  \vspace{1pt}
  \hrule
  \vspace{1pt}

	\includegraphics[width=0.45\textwidth,trim={0cm 0cm 0cm 0cm},clip]{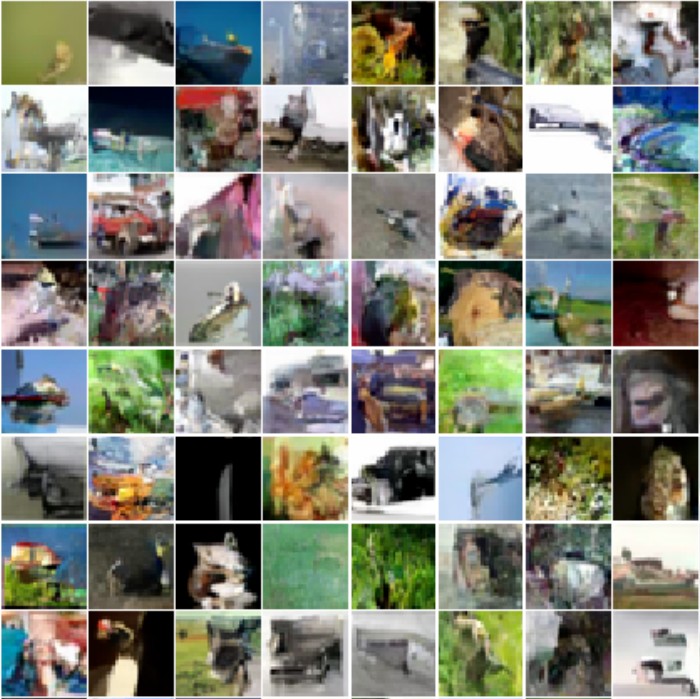}
    \caption{Samples produced by the current state-of-the-art autoregressive probabilistic image models:
            \cite{oord2016pixel}~\textit{(top)} and \cite{salimans2016pixel}~\textit{(bottom)}.}
    \label{fig:sample-pxpp}
\end{figure}

\subsection{\GRAY}

Despite great success of PixelCNN-type models, 
they are still far from producing plausible samples of complex 
natural scenes.
A visual inspection of samples produced by the current state-of-the-art models 
reveals that they typically match low-level image details well, but fail 
at capturing global image structure, such as object shapes, see 
Figure~\ref{fig:sample-pxpp} for an illustration.

We conjecture that a major reason for this is that the PixelCNN training 
objective provides too little 
incentive for the model to actually capture high-level image structure.
Concretely, the PixelCNN's loss function (the negative data log-likelihood) 
measures the amount of uncertainty of the color value of each pixel, 
conditioned on the previous ones. 
This quantity is dominated by hard-to-predict low-level 
cues, such as texture patterns, which exhibit a large uncertainty.
As a consequence, the probabilistic model is encouraged to represent such 
textures well, while visually more essential image details, such as object 
shapes, are neglected. 
We provide quantitative evidence for this claim in our experimental section.
Similar findings are also discussed in~\cite{salimans2016pixel}.

In order to tackle the aforementioned shortcoming we derive 
\GRAY, in which the auxiliary variable $\widehat{X}$ is a 4-bit 
per pixel quantized grayscale version of the original 24-bits color 
image $X$.
In combination with the factorization~\eqref{eq:likelihood}, this choice 
of auxiliary variable decouples the modeling of low and high-level image details:
the distribution $p_{\hat\theta}(\widehat X)$ (the quantized grayscale image)
contains most information about global image properties, reflecting present objects and their shapes.
The distribution $p_{\theta}(X|\widehat X)$ models missing color and texture details.
In Section~\ref{sec:experiments} we highlight the quantitative and 
qualitative effects of augmenting PixelCNN with this choice of auxiliary variable. 

\subsection{\PYRAMID}
In this section we address two further shortcomings of existing PixelCNN models.
First, the strong asymmetry of the factors in Equation~\eqref{eq:originalfactorization}: 
the top left pixel of an image is modeled unconditionally, \ie without any available 
information, while the bottom right pixel has access to the information 
of, potentially, all other pixel values. 
Nevertheless, the same network is evaluated to model either of them 
(as well as all others pixels in between). 
We conjecture that it would be beneficial if pixels were generated with
a less asymmetric usage of the image information.

Second, PixelCNNs have high computational cost for sampling images due to 
the recurrent nature of the procedure: for each generated pixel, 
the PixelCNN must invoke a convolutional neural network that is very deep, 
often in the order of a hundred convolutional layers. 
Note, that, in principle, at the sampling phase PixelCNN allows 
to cache intermediate values across consecutive PixelCNN invocations
and, thus, save considerate computational effort.
However, as reported in~\cite{ramachandran2017fastgeneration}, caching delivers minor sampling
speed gain, when only a few or a single image is generated at once,
which is a common scenario in real-life applications.

In order to alleviate the aforementioned drawbacks we propose 
a PixelCNN model in which the auxiliary variable $\widehat{X}$ corresponds 
to a twice lower resolution view of $X$, thereby decoupling the 
full image model into a pair of simpler models: creating a 
lower resolution image, and upscaling a low-res into a high-res 
image. 
The upscaling step has strongly reduced model asymmetry, because 
all pixels on the high scale have equal access to all information 
from the lower scale.
Also, by explicitly modeling the low-resolution image view we make 
it easier for the model to capture long-range image correlations, 
simply because the ``long range'' now stretches over fewer pixels.

Since the proposed auxiliary variable is an ordinary image, we can 
recursively apply the same decomposition to model the auxiliary 
variable itself. 
For example, if we apply decomposition recursively 4 times for modeling 128x128 images,
it will result in a model, which first generates an image on 8x8 resolution, and then upscales
it 4 times by a factor 2.
We call the resulting model \PYRAMID, as it resembles image pyramid decomposition.

\PYRAMIDs break the image model into a series of simpler sub-models.
Each sub-model is determined by the corresponding conditional distribution $p_{\theta}(X|\widehat{X})$
and the embedding $f_{w}(\widehat{X})$ (or just by $p_{\hat \theta}(\widehat X)$ for the lowest resolution image).
These conditional distributions model relatively simple task of producing an image,
given an embedding of the slightly lower resolution view of this image.
Thus, we hypothesize that with an appropriate embedding function (potentially modeled 
by a very deep network), the conditional distributions can be reliably modeled using 
a very light-weight network.
We then expect a significant sampling speed acceleration, because the major part of 
computational burden is redistributed to the embedding functions $f_w(\widehat X)$, 
which needs to be computed only once per pyramid layer, not once for each pixel. 

We quantify the modeling performance and sampling speed of the 
proposed multi-scale model later in Section \ref{sec:experiments}. 

\subsection{Details of model parameterization}\label{sec:parameterization}

The PixelCNN model with auxiliary variable is fully defined by factors in~\eqref{eq:likelihood},
each of which is realized by a network with the PixelCNN++
architecture. This is described in details in~\cite{salimans2016pixel}. 
The output of the PixelCNN++ is a 10-component mixture of three-dimensional logistic distributions
that is followed by a discretized likelihood function~\cite{kingma2016improving,salimans2016pixel}.
The only exception is the model for 4-bit quantized grayscale auxiliary variable $\widehat{X}$, where
output is a vector of 16 probabilities for every possible grayscale value,
followed by the standard cross-entropy loss.

The conditional PixelCNN model, $p(X|\widehat{X})$, depends on auxiliary variable $\widehat{X}$
in a fashion similar to \cite{van2016conditional,salimans2016pixel}.
It can be summarized in two steps as follows: compute an embedding of $\widehat{X}$ using a convolutional network
$f_{w}(\widehat{X})$,
and bias the convolutions of every residual block by adding the computed embedding.
We choose the architecture for modeling the embedding function, $f_w(\widehat X)$, to be
almost identical to the architecture of PixelCNN++.
The main difference is that we use only one flow of residual blocks and 
do not shift the convolutional layers outputs, because there is no need to 
impose sequential dependency structure on the pixel level.

For numeric optimization, 
we use Adam~\cite{kingma2014adam}, a variant of stochastic gradient optimization.
During training we use dropout with 0.5 rate, in a way that suggested in \cite{salimans2016pixel}.
No explicit regularization is used.
Further implementation details, such as number of layers are specified later in Section~\ref{sec:experiments}.

\section{Experiments}\label{sec:experiments}

In this section we experimentally study the 
proposed \GRAY and \PYRAMID models on natural image modeling task
and report quantitative and qualitative evaluation results. 

\subsection{\GRAY} 

\textbf{Experimental setup.} We evaluate the modeling performance of 
a \GRAY on the CIFAR-10 dataset~\cite{krizhevsky2009learning}.
It consists of 60,000 natural images of size $32 \times 32$ belonging 
to 10 categories.
The dataset is split into two parts: a training set with 50,000 images 
and a test set with 10,000 images.
We augment the training data by random horizontal image flipping.

For setting up the architectures for modeling the distributions 
$p_{\hat\theta}(\widehat{X})$, $p_{\theta}(X|\widehat{X})$
and embedding $f_w(X)$ we use the same hyperparameters as in~\cite{salimans2016pixel}.
The only exception is that for parameterizing $p_{\theta}(X|\widehat{X})$ and $f_w(\widehat X)$
we use 24 residual blocks instead of 36.

In the Adam optimizer we use an initial learning rate of 0.001, a batch size of 
64 images and an exponential learning rate decay of 0.99999 that is applied after each iteration.
We train the grayscale model $p_{\hat\theta}(\widehat{X})$ for 30 epochs
and the conditional model $p_{\theta}(X|\widehat{X})$ for 200 epochs.

\textbf{Modeling performance.} The \GRAY achieves an upper bound 
on the negative log-likelihood score of \textbf{2.98} bits-per-dimension.
This is on par with current state-of-the art models, see Table~\ref{table:nll}.
Note, that since we measure an upper bound, the actual model performance might 
be slightly better. However, in light of our and other experiments, we believe 
small differences in this score to be of minor importance, as the log-likelihood 
does not seem to correlate well with visual quality in this regime. 

\begin{figure*}
    \center
	\includegraphics[width=1.0\textwidth,trim={0 1.4cm 4.04cm 0},clip]{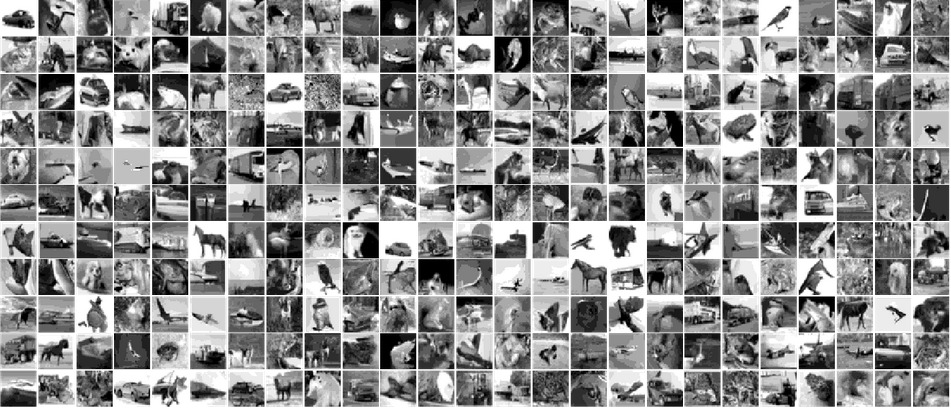}

  \vspace{1mm}

  \includegraphics[width=1.0\textwidth,trim={0 1.4cm 4.04cm 0},clip]{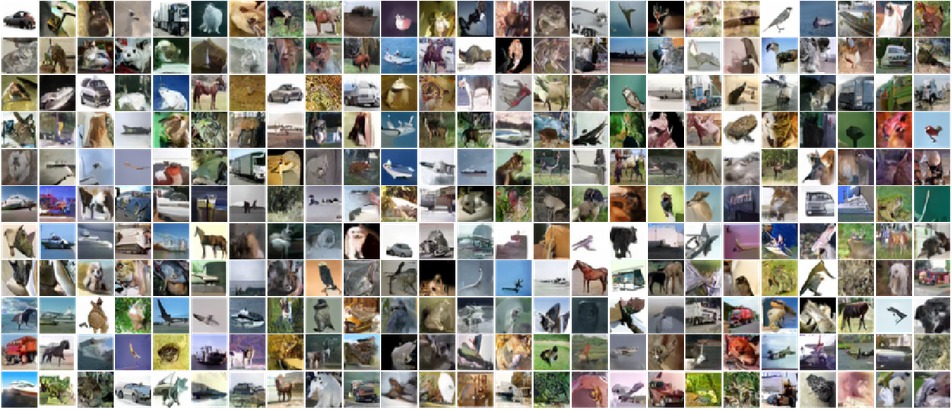}
  \caption{Random quantized grayscale samples from $p(\widehat{X})$ \textit{(top)} and
           corresponding image samples from $p(X|\widehat{X})$ \textit{(bottom)}. The 
           grayscale samples show several recognizable objects, which are subsequently
           also present in the color version.}
    \label{fig:samples}
\end{figure*}

In Figure~\ref{fig:samples} we present random samples produced 
by the \GRAY model, demonstrating grayscale samples from 
$p_{\hat\theta}(\widehat{X})$ and resulting colored samples from $p_{\theta}(X|\widehat{X})$.
We observe that the produced samples are highly diverse, and, unlike samples 
from previously proposed probabilistic autoregressive models, often exhibit a 
strongly coherent global structure, resembling highly complex objects, such as 
cars, dogs, horses, \etc.

Given the high quality of the samples, one might be worried if possibly 
the grayscale model, $p_{\hat\theta}(\widehat X)$, had overfit the training data. 
We observe that training and test loss of $p_{\hat\theta}(\widehat{X})$ are 
very close to each other, namely 0.442 and 0.459 of bits-per-dimension, 
which speaks against significant overfitting. 
Note also, that it is not clear if an overfitted model would automatically produce good
samples.
For instance, as reported in~\cite{salimans2016pixel}, severe overfitting of the PixelCNN++
model does not lead to a high perceptual quality of sampled images.

\textbf{Discussion.} By explicitly emphasizing the modeling of high-level 
image structures in the \GRAY, we achieve significantly better visual 
quality of the produced samples.
Additionally, the \GRAY offers interesting insights into the image modeling task. 
As the objective we minimize for training is a sum of two scores, we can individually examine 
the performance of auxiliary variable model $p(\widehat{X})$ and the conditional model $p(X|\widehat{X})$.
The trained conditional model achieves $p(X|\widehat{X})$ a negative log-likelihood score of \textbf{2.52} 
bits-per-dimension, while $p(\widehat{X})$ achieves a score of \textbf{0.459} bits-per-dimension on the 
CIFAR-10 test set. In other words: high-level image properties, despite being harder to model for the 
network, contribute only a small fraction of the uncertainty score to the the overall log-likelihood. 
We take this as indication that if low-level and high-level image details are modeled by one 
model, then at training time low-level image uncertainty will dominate the training objective.
We believe that this is this the key reason why previously proposed PixelCNN models failed
to produce globally coherent samples of natural images.
Importantly, this problem does not appear in the \GRAY, because global and local image models
do not share parameters and, thus, do not interfere with each other at training phase.

\begin{figure*}
    \center
	\includegraphics[width=0.25\textwidth]{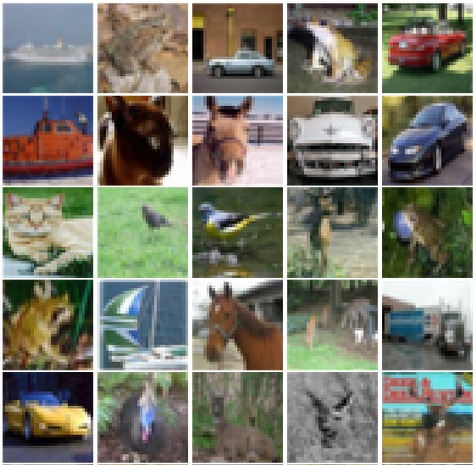}
    \quad
	\includegraphics[width=0.25\textwidth]{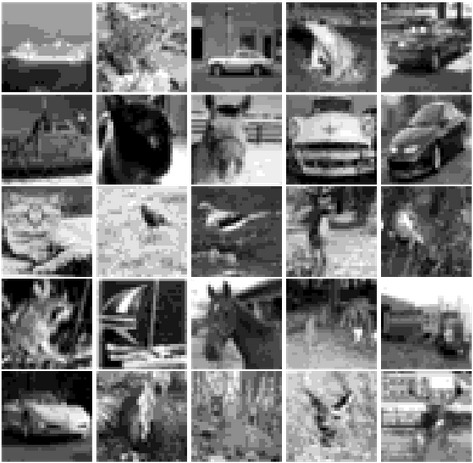}
    \quad
	\includegraphics[width=0.25\textwidth]{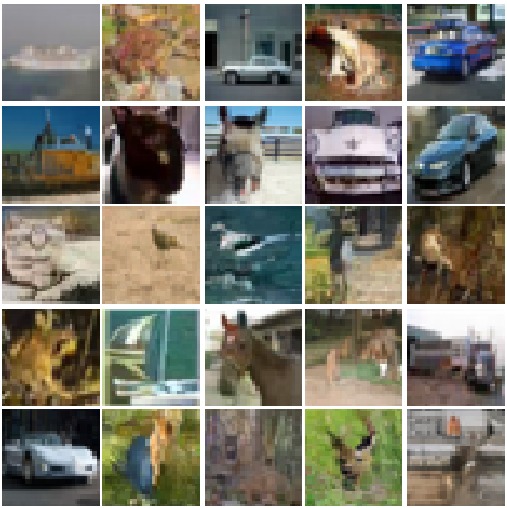}
    \caption{CIFAR-10 images in original color \emph{(left)} and quantized to 4-bit grayscale \emph{(center)}.
     Images sampled from our conditional model $p(X|\widehat{X})$, using the grayscale CIFAR images as auxiliary 
     variables \emph{(right)}. The images produced by our model are visually as plausible as the original ones.}
    \label{fig:coloring}
\end{figure*}

An alternative explanation for the differences in log-likelihood scores would be 
that PixelCNN-type models are actually not very good at modeling low-level image input. 
Additional experiments that we performed show that this is not the case: 
we applied the learned conditional model $p(X|\widehat{X})$ to 4-bit grayscale images
obtained by quantizing real images of the CIFAR-10 test set. Figure~\ref{fig:coloring} 
compares the resulting colorized samples with the corresponding original images,
showing that the samples produced by our conditional model are of visual quality comparable
to the original images.
This suggests that in order to produce even better image samples, 
mainly improved models for $p_{\hat\theta}(\widehat{X})$ are required. 

\begin{table}[t]
        \center
	\resizebox{0.5\textwidth}{!}{
        \begin{tabular}{|lc|}
        \hline
	Model & Bits per dim. \\
        \hline
      Deep Diffusion \cite{sohl2015deep} & $\le$ 5.40 \\
                NICE \cite{dinh2014nice} & 4.48 \\
                DRAW \cite{gregor2015draw} & $\le$ 4.13 \\
           Deep GMMs \cite{van2014factoring} & 4.00 \\
           Conv Draw \cite{gregor2016towards} & $\le$ 3.58 \\
            Real NVP \cite{dinh2016density} & 3.49 \\
  Matnet + AR \cite{bachman2016architecture} & $\le$ 3.24 \\
            PixelCNN \cite{oord2016pixel} & 3.14 \\
        VAE with IAF \cite{kingma2016improving} & $\le$ 3.11 \\
      Gated PixelCNN \cite{van2016conditional} & 3.03 \\
            PixelRNN \cite{oord2016pixel} & 3.00 \\
  \textbf{\GRAY} (this paper) & $\le$ 2.98 \\
 DenseNet VLAE \cite{chen2016variational} & $\le$ 2.95 \\
	PixelCNN++ \cite{salimans2016pixel} & 2.92 \\
        \hline
        \end{tabular}
	}
    \caption{The negative log-likelihood of the different models for the CIFAR-10 \textbf{test set}
             measured as bits-per-dimension.}
	\label{table:nll}
\end{table}

\subsection{\PYRAMID.} 

\textbf{Experimental setup.} We evaluate the \PYRAMID on 
the task of modeling face images. We rely on the \emph{aligned$\&$cropped CelebA} 
dataset~\cite{liu2015faceattributes} that contains approximately 
200,000 images of size 218x178.
In order to focus on human faces and not background, we preprocess all images in the dataset by applying a 
fixed 128x128 crop (left margin: 25 pixels, right margin: 
25 pixel, top margin: 50, bottom margin: 40 pixels).
We use a random 95\% subset of all images as training set 
and the remaining images as a test set.

For the \PYRAMID we apply the auxiliary variable decomposition 4 times.
This results in a sequences of probabilistic models, where the first model generates 
faces in 8$\times$8 resolution. 
%
%
We use a PixelCNN++ architecture without down or up-sampling layers with 
only 3 residual blocks to model the distributions $p_{\theta}(\widehat{X})$ and $p_{\hat{\theta}}(X|\widehat{X})$
for all scales.
For the embedding $f_w(\widehat{X})$ we use a PixelCNN++ architecture with
15 residual blocks with downsampling layer after the residual block 
number 3 and upsampling layers after the residual blocks number 9 and 12.
For all convolutional layers we set the number of filters to 100.

In the Adam optimizer we use an initial learning rate 0.001, a batch 
size of 16 and a learning rate decay of 0.999995.
We train the model for 60 epochs.

\begin{table}[t]
	\center
        \begin{tabular}{|l||c|c|c|c|c|}
        \hline
		Res. & 8$\times$8 & 16$\times$16 & 32$\times$32  & 64$\times$64 & 128$\times$128 \\
		Bpd  & 4.58 & 4.30 & 3.33 & 2.61 & 1.52 \\
	\hline
	\end{tabular}
	\caption{Bits-per-dimension (\emph{Bpd}) achieved by \PYRAMID on the test images
	         from the \emph{CelebA} dataset for various resolutions (\emph{Res.}).}
	\label{table:CelebA}
\end{table}

\textbf{Modeling performance.}
We present a quantitative evaluation of \PYRAMID in Table~\ref{table:CelebA}.
We evaluate the performance of our model on different output resolutions,
observing that bits-per-dimension score is smaller for higher resolutions,
as pixel values are more correlated and, thus, easier to predict.
As an additional check, we also train and evaluate \PYRAMID model on
the CIFAR dataset, achieving a competitive score of 3.32 bits-per-dimension.

\begin{figure*}
    \center
	\includegraphics[width=.98\textwidth]{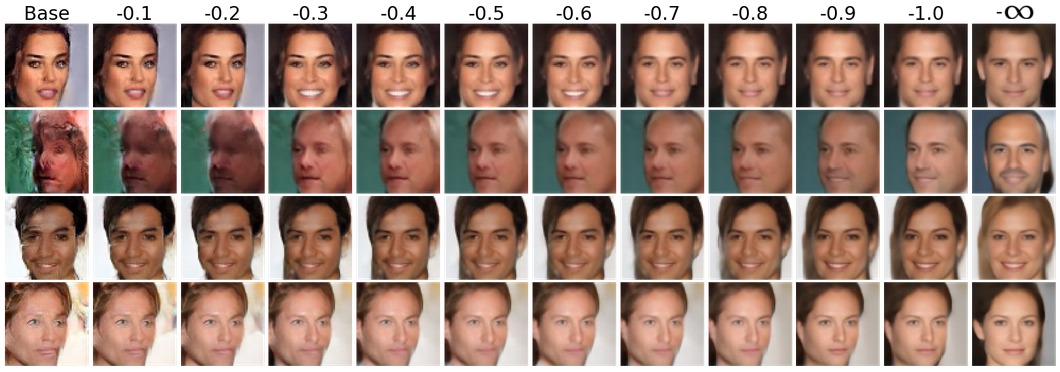}
	\caption{Effect of the variance reduction. Numbers on top of each column indicates the amount of reduction
             in the predicted log-variance of the mixture components. The last column corresponds to \SAMPLING.}
    \label{fig:face-variance}
\end{figure*}

Before demonstrating and discussing face samples produced by our model 
we make an observation regarding the PixelCNN's sampling procedure.
Recall that the output of the \PYRAMID is a mixture of logistic 
distributions.
We observe an intriguing effect related to the mixture representation 
of the predicted pixel distributions for face images:
the perceptual quality of sampled faces substantially increases if 
we artificially reduce the predicted variance of the mixture components.
We illustrate this effect in Figure~\ref{fig:face-variance}, where 
we alter the variance by subtracting constants from a fixed set 
of $\{0.0, 0.1, \dots, 1.0\}$ from the predicted log-variance of 
the mixture components.
Inspired by this observation we propose an alternative sampling procedure: 
for each pixel, we randomly sample one of the logistic components based 
on their weight in the predicted mixture. Then, we use the mode of this 
component as sampled pixel value, instead of performing a second random 
sampling step.
This sampling procedure can be seen as a hybrid of 
probabilistic sampling and maximum a posteriori (MAP) prediction. 

\begin{figure*}
    \center
	\includegraphics[width=1.0\textwidth]{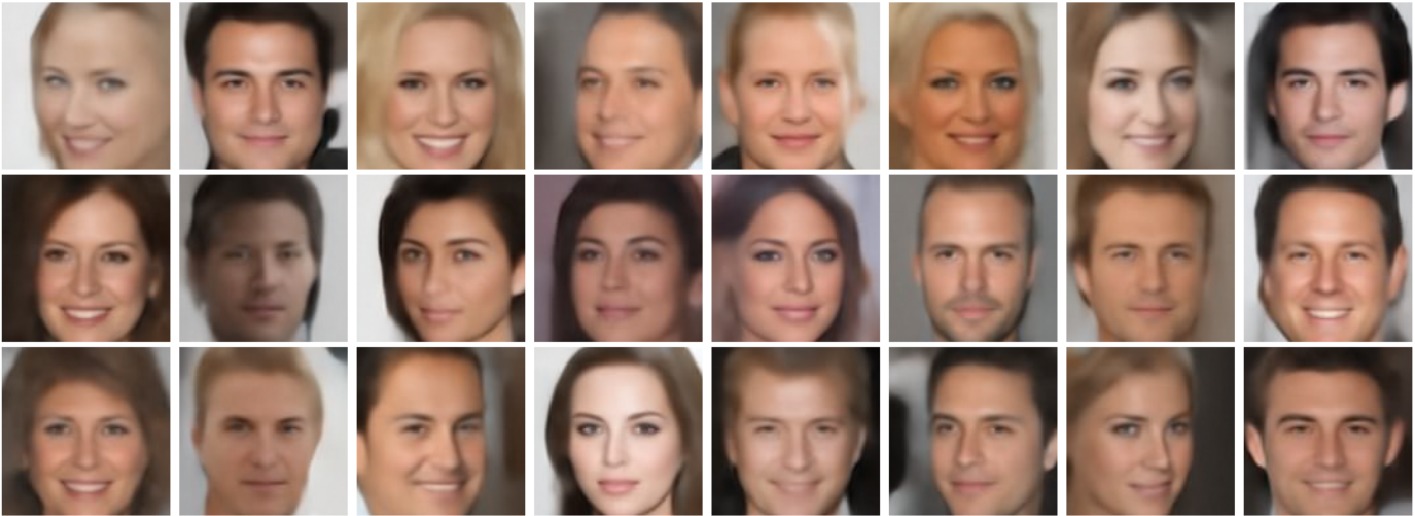}
	\caption{Images sampled from the \PYRAMID by \SAMPLING. The generated faces are of very high quality,
    many being close to photorealistic. At the same time, the set of sample is diverse in terms of the depicted 
    gender, skin color and head pose.}
    \label{fig:face-samples}
\end{figure*}

\begin{figure*}
    \center
	\includegraphics[width=1.0\textwidth]{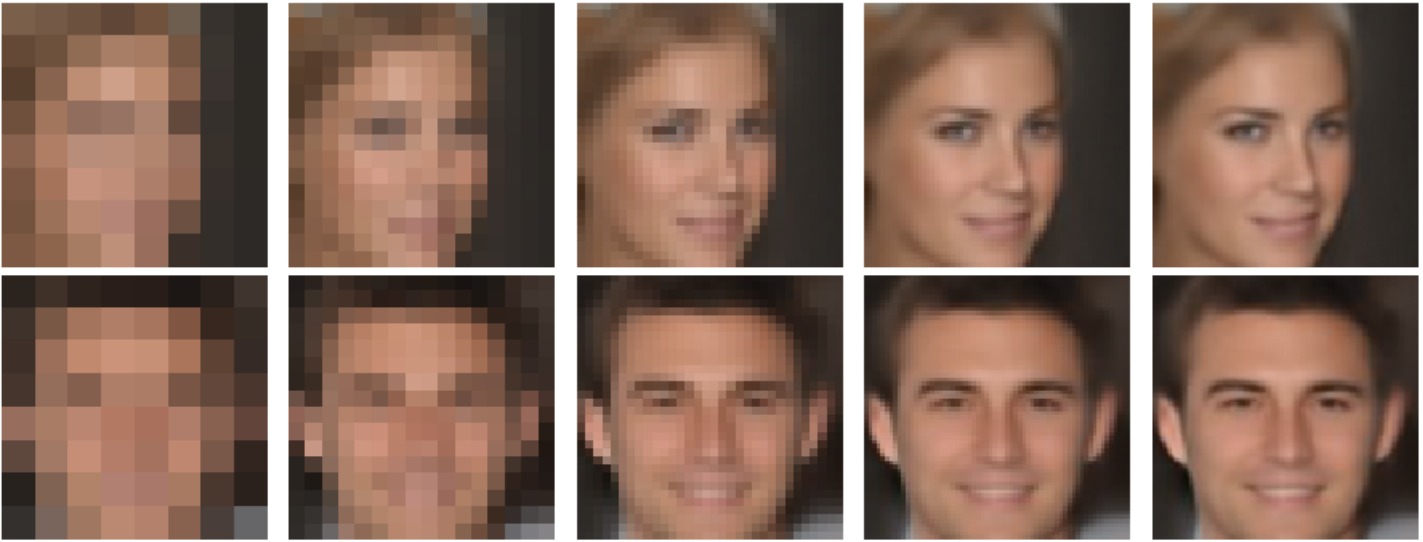}
	\caption{Visualization of the \PYRAMID sampling process. Faces are first generated on a small, 8x8, resolution
             and then are gradually upsampled until reaching the desired 128x128 resolution.}
    \label{fig:face-upsampling}
\end{figure*}

Figure~\ref{fig:face-samples} shows further samples obtained by such 
\emph{\SAMPLING}. 
The produced images have very high perceptual quality, with some generated 
faces appearing almost photo-realistic.
The complete multi-scale sampling mechanism of the \PYRAMID, from 
8$\times$8 to 128$\times$128 images, is demonstrated in Figure~\ref{fig:face-upsampling}. 

\textbf{Discussion.} First, we observe that, despite the very high 
resolution of modeled images, the produced samples capture global human 
face characteristics, such as arrangement of face elements and global 
symmetries. At the same time, the set of samples is diverse, containing 
male as well as female faces, different hair and skin colors as well 
as facial expressions and head poses. 
Second, we emphasize that by properly decomposing the model we are 
able to scale the \PYRAMID to produce samples with very high 
resolution of 128x128.
As discussed previously in Section~\ref{sec:method}, this results from 
the fact that our decomposition allows to parametrize autoregressive 
parts of the image model by a light-weight architecture.
Concretely, on an NVidia TitanX GPU, our \PYRAMID without 
caching optimizations requires approximately \textbf{0.004} seconds 
on average to generate one image pixel, while a PixelCNN++ even with 
recently suggested caching optimizations requires roughly \textbf{0.05} 
seconds for the same task.
If we add caching optimizations to our model, we expect its speed to
improve even further.

\vspace{-1mm}
\section{Conclusion}

In this paper we presented \GRAY and \PYRAMID, an improved autoregressive 
probabilistic techniques that incorporate auxiliary variables.
We derived two generative image models that exploit different image 
views as auxiliary variables and address known limitations of existing 
PixelCNN models.
The use of quantized grayscale images as auxiliary variables 
resulted in a model that captures global structure
of complex natural images and produces globally coherent 
samples.
With multi-scale image views as auxiliary variable, the model was able to 
efficiently produce realistic high-resolution images of
human faces.
Note, that these improvements are complementary and we plan to combine 
them in a future work.

Furthermore, we gained interesting insights into the image modeling problem.
First, our experiments suggest that texture and other low-level
image information distract probabilistic models from focusing 
on more essential high-level image information, such as object shapes.
Thus, it is beneficial to decouple the modeling of low and high-level 
image details. 
Second, we demonstrate that multi-scale image model, even with very 
shallow PixelCNN architectures, can accurately model high-resolution 
face images.

\textbf{Acknowledgments.} We thank Tim Salimans for spotting a mistake in our preliminary 
\emph{arXiv} manuscript. 
This work was funded by the European Research Council under 
the European Unions Seventh Framework Programme (FP7/2007-2013)/ERC 
grant agreement no 308036.

\bibliography{icml}
\bibliographystyle{icml2017}

\end{document}